\documentclass[runningheads]{llncs}
\usepackage{amsmath,amssymb,amsfonts}
\usepackage{multirow}
\usepackage{rotating}
\usepackage{makecell}
\usepackage{algorithmic}
\usepackage{textcomp}
\usepackage{subfigure}
\usepackage[T1]{fontenc}
\newcommand{\figref}[1]{Fig.~\ref{#1}}
\newcommand{\tabref}[1]{Tabel~\ref{#1}}

\usepackage{overpic}
\usepackage{xcolor}
\usepackage{bbding} 
\usepackage{graphicx}
\usepackage[colorlinks,linkcolor=red, anchorcolor=blue, citecolor=blue, urlcolor=magenta]{hyperref}
\usepackage{bbm}
\usepackage{algorithmic}
\usepackage{textcomp}
\usepackage[misc]{ifsym}
\usepackage{rotating,soul}
\usepackage{gensymb}
\begin{document}
\title{Unsupervised Domain Adaptive Fundus Image Segmentation with Category-level Regularization}
%
%
\author{Wei Feng\inst{1,2,3} \and 
Lin Wang\inst{1,2,3}\and 
Lie Ju\inst{1,2,3}\and 
Xin Zhao\inst{4} \and 
Xin Wang\inst{4} \and 
Xiaoyu Shi\inst{4} \and 
Zongyuan Ge\inst{1,2,3}$^{(\textrm{\Letter})}$} 
\index{Feng, Wei}
\index{Wang, Lin}
\index{Ju, Lie}
\index{Zhao, Xin}
\index{Wang, Xin}
\index{Shi, Xiaoyu}
\index{Ge, Zongyuan}
%
\institute{Monash Medical AI Group, Monash University, Melbourne, Australia\and
 Airdoc Monash Research Centre, Monash University, Clayton, Australia \and 
Monash eResearch Center, Monash University, Clayton, Australia \and
Airdoc LLC, Beijing, China \\
\email{wf02429@gmail.com}, \email{zongyuan.ge@monash.edu}\\
\url{https://www.monash.edu/mmai-group} \\
}
%
\maketitle              
\begin{abstract}
Existing unsupervised domain adaptation methods based on adversarial learning have achieved good performance in several medical imaging tasks. However, these methods focus only on global distribution adaptation and ignore distribution constraints at the category level, which would lead to sub-optimal adaptation performance. This paper presents an unsupervised domain adaptation framework based on category-level regularization that regularizes the category distribution from three perspectives. Specifically, 
for inter-domain category regularization, an adaptive prototype alignment module is proposed to align feature prototypes of the same category in the source and target domains.
In addition, for intra-domain category regularization, we tailored a regularization technique for the source and target domains, respectively. In the source domain, a prototype-guided discriminative loss is proposed to learn more discriminative feature representations by enforcing intra-class compactness and inter-class separability, and as a complement to traditional supervised loss.
In the target domain, an augmented consistency category regularization loss is proposed to force the model to produce consistent predictions for augmented/unaugmented target images, which encourages semantically similar regions to be given the same label. Extensive experiments on two publicly fundus datasets show that the proposed approach significantly outperforms other state-of-the-art comparison algorithms\footnote[1]{Our code is available at \url{https://github.com/fengweie/UDA_CLR}.}. 

\end{abstract}
\section{Introduction}
Recently deep neural networks have dominated several medical image analysis tasks and have achieved good performance \cite{wu2018multiscale,ju2022improving,hang2020local}. 
However, a well-trained model usually underperform when tested directly on an unseen dataset due to domain shift \cite{wang2019boundary}.
In clinical practice, this phenomenon is prevalent and remains unresolved. To this end, domain adaptation strategies have received a lot of attention, aiming to transfer knowledge from a label-rich source domain to a label-rare target domain. Recent adversarial training-based domain adaptation methods have shown promising performance, focusing mainly on global distribution adaptation at the input space \cite{zhu2017unpaired}, feature space \cite{dou2018unsupervised} or output space \cite{wang2019patch}. Despite the significant performance gains achieved, they all ignore the category distribution constraints. This may result in a situation where although global distribution differences between domains have been reduced, the pixel features of different categories in the target domain are not well separated. This is because some categories are similar to others in terms of appearance and texture. There have been several studies try to address this issue. For example, Liu et al. \cite{liu2021bapa} proposed a prototype alignment loss to reduce the mismatch between the source and target domains in the feature space. Xie et al. \cite{xie2018learning} proposed a semantic alignment loss to learn semantic feature representations by aligning the category centres of the labelled source domain and the category centres of the pseudo-labelled target domain. However, the shortcoming of these methods is that there is no explicit constraint on the distance between different category features, resulting in categories that look similar in the source domain also being distributed similarly in the target domain, which would potentially lead to incorrect 
prediction results, especially in edge regions and low-contrast regions.

In this paper, we propose an unsupervised domain adaptation framework based on category-level regularization to accurately segment the optic disc and cup from fundus images. We perform category regularization from both intra-domain and inter-domain perspectives. Specifically, for intra-domain category regularization,
on the source domain side, we first propose a prototype-guided discriminative loss to enhance the separability of inter-class distributions and the compactness of intra-class distributions, thus learning more discriminative feature representations; on the target domain side, we propose an augmented consistency-based category regularization loss to constrain the model to produce consistent predictions for perturbed and unperturbed target images, thus encouraging semantically similar regions to have the same labels.
For inter-domain category regularization, we propose an adaptive prototype alignment 
module to ensure that pixels from the same class but different domains can be mapped nearby in the feature space. Experiment results on two public fundus datasets and ablation studies demonstrate the effectiveness of our approach. 

\section{METHODOLOGY}
\begin{figure*}[t]
    \centering
    \includegraphics[width=\linewidth]{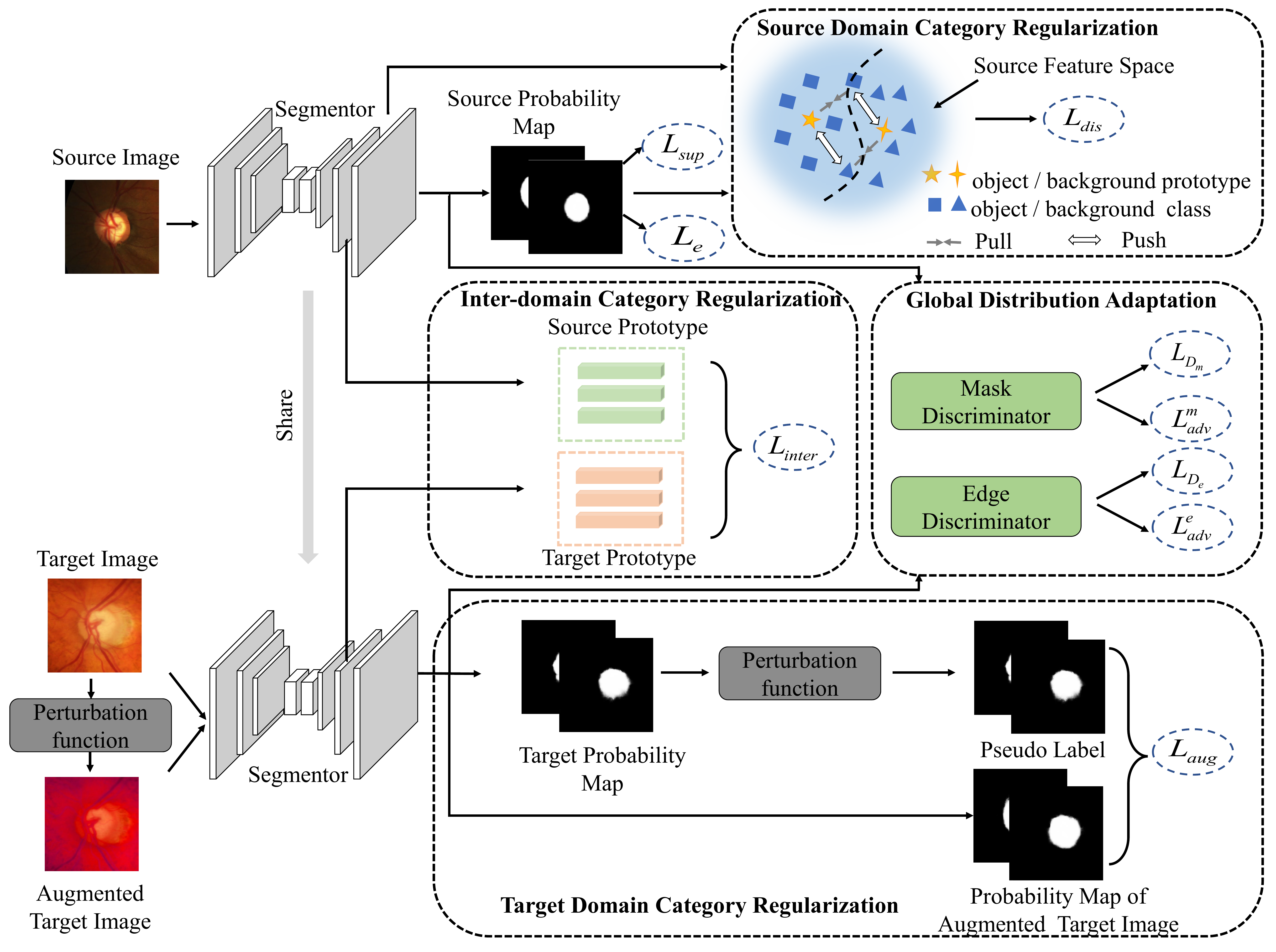}
    \caption{Overview of an unsupervised domain adaptation framework based on category-level regularization. We first align global distributions between domains by adversarial learning. Then we perform fine-grained level category distribution adaptation from three perspectives: source domain, target domain and inter-domain, via three category regularization methods.}
    \label{fig.2}
\end{figure*}
Fig.~\ref{fig.2} shows an overview of our proposed unsupervised domain adaptation framework based on category-level regularization. It consists of three main components, prototype-guided source domain category regularization, augmented consistency-based target domain category regularization, and inter-domain category regularization, performing category-level regularization from different perspectives.

\subsection{Inter-domain Category Regularization}
In an typical unsupervised domain adaptation (UDA) setting, we are given a source domain image set $\{x_i^s\}_{i=1}^{N_s}$ and its corresponding pixel-wise annotation $\{y_i^s\}_{i=1}^{N_s}$, and a target domain image set $\{x_i^t\}_{i=1}^{N_t}$ without annotation. 
To regularize the category distributions between domains, we propose an adaptive prototype alignment module that aligns prototypes of pixels of the same category in the labelled source domain and the pseudo-labelled target domain, thus guaranteeing that features of the same category in different domains are mapped nearby.

Specifically, we feed the target images $x^t$ into the segmentation model $G$ to obtain the pseudo-label $\widehat y^t=\mathbbm{1}[p^t \geqslant \beta]$, where $p^t$ is the predicted probability and $\mathbbm{1}$ is the indicator function. $\beta$ is a probability threshold. We denote the feature map before the last convolution layer as $h^{t}$. Based on $\widehat y^t$ and $h^{t}$, we can obtain the object prototype $f_{obj}^{t}$ of the target images as:
\begin{equation}\label{equ:trgpro}
f_{obj}^{t}=\frac{1}{N_{obj}} \sum_{k} \mathbbm{1} \left(\widehat y_{k}^{t}=1\right) h_{k}^{t},\quad \\
\end{equation}
where $k$ represent the pixel index,${N_{obj}}$ is
the number of pixels of the object class.

However, due to differences in distribution between domains, pseudo labels of target images may contain some incorrect predictions and these false pseudo labels
 may affect the prototype computation \cite{liu2021bapa}. Inspired by \cite{yu2019uncertainty}, we introduce an uncertainty-guided noise-aware module to filter out those unreliable pseudo labels. We estimate the uncertainty of the pseudo labels based on the Monte Carlo dropout method \cite{gal2016dropout}. Specifically, we enable dropout and perform $M$ stochastic forward inferences to obtain $M$ predicted outputs $\{p_m^t\}_{m=1}^M$. We are then able to obtain the uncertainty estimate for each pixel based on the standard deviation $S=std(\{p_m^t\}_{m=1}^M)$ of the multiple model predictions. We then filter out the pseudo labels of those unreliable pixels by setting an uncertainty threshold $\xi$ to avoid their impact on the prototype computation. 
 The object prototype $f_{obj}^{s}$ of the source domain is calculated in a similar way\footnote[2]{Note that since source domain annotation information is available, we use ground truth labels to compute the source domain prototypes.}. Then the inter-domain category regularization loss can be formulated as:
\begin{equation}\label{equ:proloss}
L_{inter}=D(f^{s}_{obj}-f^{t}_{obj}), 
\end{equation}
where $D$ is the distance function, we use the Euclidean distance. Note that we only align the prototypes of object class between domains, as object class has more domain shared features than background class \cite{zheng2020cross}.
\subsection{Intra-domain Category Regularization}
In order to further regularize the distributions of the different categories in the feature space, we perform intra-domain category regularization to learn discriminative feature representations by using the category information within the source and target domains, which also works as a complement of inter-domain category regularization.


\subsubsection{Source Domain Category Regularization}
On the source domain side, we propose a prototype-guided discriminative loss to regularize the category distribution. Specifically, we use the category feature prototype to provide supervised signals, explicitly constraining pixel features to be closer to their corresponding category prototypes, while being farther away from other category prototypes. The prototype-guided discriminative loss is formulated as:
\begin{equation}\label{equ:dis}
\begin{aligned}
L_{d i s}=&\sum_{k}\mathbbm{1}\left(y_{k}^{s}=1\right) \max \left(\left\|h_{k}^{s}-f_{obj}^{s}\right\|-\left\|h_{k}^{s}-f_{bg}^{s}\right\|+\delta, 0\right) \\
&+\sum_{k}\mathbbm{1}\left(y_{k}^{s}=0\right) \max \left(\left\|h_{k}^{s}-f_{bg}^{s}\right\|-\left\|h_{k}^{s}-f_{obj}^{s}\right\|+\delta, 0\right),
\end{aligned}
\end{equation}
where $\delta$ is a predefined distance margin. $f_{bg}^{s}=\frac{1}{N_{bg}} \sum_{k} \mathbbm{1}
\left(y_{k}^{s}=0\right) h_{k}^{s}$ is the prototype of the background class. $h^{s}$ is the pixel-wise deep feature of the source domain images. This loss would be 0 when the distance between each pixel feature and its corresponding prototype is less than its distance from other classes of prototypes by a margin $\delta$.
\subsubsection{Target Domain Category Regularization}
In the target domain, since we do not have access to the ground truth labels and therefore the discriminative loss can not be applied as the same way as in the source domain. To perform category-level feature regularization, inspired by the dominant consistency training strategy in semi-supervised learning \cite{wang2021deep}, we propose an augmented consistency-based regularization method that constrains the predictions of the augmented target images to be consistent with the pseudo labels of the original target images, which encourages semantically similar parts of the target images to have the same labels and thus regularize the category-level feature distributions. 

Specifically, we apply a perturbation function to $(x^{t},\widehat y^t)$ to generate a perturbed pair $(x^{pert},\hat{y}^{pert})$.The augmented consistency loss can be formulated as:
\begin{equation}\label{equ:aug}
L_{aug}=-\sum_{k} \mathbbm{1}\left(S_{k}<\mu\right) \ell\left(G(x_{k}^{pert}), \hat{y}_{k}^{pert}\right),
\end{equation}
where $\ell(\cdot)$ is the cross-entropy loss. Note that here we only calculate the augmented consistency loss for those pseudo-labelled pixels for which the uncertainty estimate $S_{k}$ is less than a threshold $\mu$ to avoid error accumulation due to incorrect pseudo labels \cite{sohn2020fixmatch}.
\subsection{Training Procedure}
In addition to category-level regularization, we also perform global distribution alignment by adversarial learning. Following \cite{wang2019boundary}, we build two discriminators, $D_{m}$ and $D_{e}$, to align the predicted probability distribution $(p^{s}_{m},p^{t}_{m})$ and the edge structure distribution $(p^{s}_{e},p^{t}_{e})$ of the source and target domains respectively. At the same time the training goal of the segmentation model is to learn domain invariant features to deceive the discriminators. In summary, the training objective of the segmentation network can be formulated as:
\begin{equation}\label{equ:segmentation}
\begin{aligned}
L_{total} & = L_{sup} + L_{e}+\lambda_{1} L_{d}+ \lambda_{2} L_{inter}+ \lambda_{3} L_{dis}+ \lambda_{4} L_{aug}\\
L_{e}&= \sum_{k}\left(y^{s}_{e,k}-p^{s}_{e,k}\right)^{2}\\
L_{d}&= L_{adv}^{m}+L_{adv}^{e}=\frac{1}{N_{t}} \sum_{i=1}^{N_{t}} L_{D}\left(p^{t}_{m,i}, 1\right)+\frac{1}{N_{t}} \sum_{i=1}^{N_{t}} L_{D}\left(p^{t}_{e,i}, 1\right),
\end{aligned}
\end{equation}
where $L_{sup}$ is the supervised loss on the labelled source domain image. $L_{e}$ and $L_{d}$ are the edge regression loss and the adversarial loss respectively, $L_{D}$ is the binary cross-entropy loss. $y^{s}_{e}$ is the edge ground truth labels. $\lambda_{1},\lambda_{2},\lambda_{3},\lambda_{4}$ are balance coefficients. 

The training objectives of the two discriminators are:
\begin{equation}\label{equ:discriminator}
\begin{aligned}
L_{D_m}&=\frac{1}{N_s} \sum_{i=1}^{N_{s}} L_{D}\left(p^{s}_{m,i}, 1\right)+\frac{1}{N_t} \sum_{i=1}^{N_{t}} L_{D}\left(p^{t}_{m,i}, 0\right)\\
L_{D_e}&=\frac{1}{N_s} \sum_{i=1}^{N_{s}} L_{D}\left(p^{s}_{e,i}, 1\right)+\frac{1}{N_t} \sum_{i=1}^{N_{t}} L_{D}\left(p^{t}_{e,i}, 0\right),
\end{aligned}
\end{equation}
where $L_{D_m}$ and $L_{D_e}$ are the adversarial loss of the mask discriminator and the adversarial loss of the edge discriminator, respectively.
\section{Experiments}
\label{sec:experiments}
\subsubsection{Dataset and evaluation metric.}
In order to evaluate the proposed method, we use three datasets: the training part of the REFUGE challenge\footnote[3]{\color{magenta}{\url{https://refuge.grand-challenge.org/}}} \cite{orlando2020refuge}, RIMONE-r3 \cite{fumero2011rim} and Drishti-GS \cite{sivaswamy2015comprehensive}. Following \cite{wang2019boundary}, we choose the REFUGE challenge as the source domain and RIMONE-r3 and Drishti-GS as the target domains, respectively. The training set of the REFUGE challenge contains 400 images with annotations, and the RIMONE-r3 and Drishti-GS contain 99/60 and 50/51 training/testing images respectively. Following \cite{wang2019boundary}, we crop a 512x512 optic disc region as input of the model. In addition, we use the commonly used Dice coefficient to evaluate the segmentation performance of the optic disc and cup \cite{wang2019boundary}.
\subsubsection{Implementation details.}
We use the MobileNetV2 modified Deeplabv3+ \cite{chen2018encoder} network as the segmentation backbone \cite{wang2019boundary}. The Adam algorithm is used to optimize the segmentation model and SGD algorithm is used to optimize the two discriminators \cite{wang2019boundary}. The learning rate of the segmentation network is set as $1e-3$ and divided by 0.2 every 100 epochs and we train a total of 500 epochs. The learning rate of the two discriminators is set as $2e-5$.
  The probability threshold $\beta$ is set as 0.75 \cite{wang2019boundary}. In the uncertainty estimation part, we perform 10 stochastic forward passes, and the uncertainty threshold $\mu$ is set as 0.05 \cite{chen2021source}. We empirically set the distance margin $\delta$ to 0.01 and found that it worked well on different datasets. The loss balance coefficients $\lambda_{1},\lambda_{2},\lambda_{3},\lambda_{4}$ are set as 0.01, 0.01, 0.01,0.01. For the perturbation function, we use the perturbation function used in \cite{chen2020simple}, which includes: color jittering and gaussian blur. 
  We use the feature map of the previous layer of the last convolutional layer to calculate the prototype.
  All experiments are performed using the Pytorch framework and 8 RTX 3090 GPUs.
\begin{table*}[t!]
\centering
    \caption{Comparison of different methods on the target domain datasets}
     \label{tab:compare}
     \setlength{\tabcolsep}{4mm}
\begin{tabular}{l|cc|cc}
\hline
\multirow{2}{*}{Method} & \multicolumn{2}{c|}{RIM-ONE-r3 \cite{fumero2011rim}}   & \multicolumn{2}{c}{Drishti-GS \cite{sivaswamy2015comprehensive}}   \\ \cline{2-5} 
                        & \multicolumn{1}{c|}{Dice disc} & Dice cup & \multicolumn{1}{c|}{Dice disc} & Dice cup \\ \hline
Oracle                  & \multicolumn{1}{c|}{0.968}     & 0.856    & \multicolumn{1}{c|}{0.974}     & 0.901    \\ \hline
Baseline                & \multicolumn{1}{c|}{0.779}     & 0.744    & \multicolumn{1}{c|}{0.944}     & 0.836    \\ \hline
TD-GAN   \cite{zhang2018task}               & \multicolumn{1}{c|}{0.853}     & 0.728    & \multicolumn{1}{c|}{0.924}     & 0.747    \\ \hline
Hoffman et al.  \cite{hoffman2016fcns}         & \multicolumn{1}{c|}{0.852}     & 0.755    & \multicolumn{1}{c|}{0.959}     & 0.851    \\ \hline
Javanmardi et al. \cite{javanmardi2018domain}     & \multicolumn{1}{c|}{0.853}     & 0.779    & \multicolumn{1}{c|}{0.961}     & 0.849    \\ \hline
OSAL-pixel \cite{wang2019patch}              & \multicolumn{1}{c|}{0.854}     & 0.778    & \multicolumn{1}{c|}{0.962}     & 0.851    \\ \hline
pOSAL \cite{wang2019patch}                   & \multicolumn{1}{c|}{0.865}     & 0.787    & \multicolumn{1}{c|}{0.965}     & 0.858    \\ \hline
BEAL \cite{wang2019boundary}                   & \multicolumn{1}{c|}{0.898}     & 0.810    & \multicolumn{1}{c|}{0.961}     & 0.862    \\ \hline
\textbf{Ours}           & \multicolumn{1}{c|}{\textbf{0.905}} & \textbf{0.841} & \multicolumn{1}{c|}{\textbf{0.966}} & \textbf{0.892} \\ \hline
\end{tabular}
\end{table*}
\subsubsection{Comparison with state-of-the-art methods.}
We compare the proposed method with Baseline method (without adaptation), fully supervised methods (Oracle), and several state-of-the-art unsupervised domain adaptation algorithms, including TD-GAN \cite{zhang2018task}, high-level alignment \cite{hoffman2016fcns}, output space adaptation \cite{javanmardi2018domain,wang2019patch}, BEAL \cite{wang2019boundary}. As can be seen from the experimental results in \tabref{tab:compare}, the proposed method achieves significant performance gains on both datasets, especially for the segmentation of the optic cup. Compared to the best comparison algorithm BEAL, our method achieves 3.1\% and 3\% Dice improvement for the optic cup segmentation on the RIM-ONE-r3 and Drishti-GS datasets, respectively. Furthermore, the segmentation performance of our method is very close to that of fully supervised performance. This indicates that our method is able to achieve good performance in scenarios with varying degrees of domain shift.

\begin{figure*}[t]
    \centering
            \includegraphics[width=0.9\linewidth]{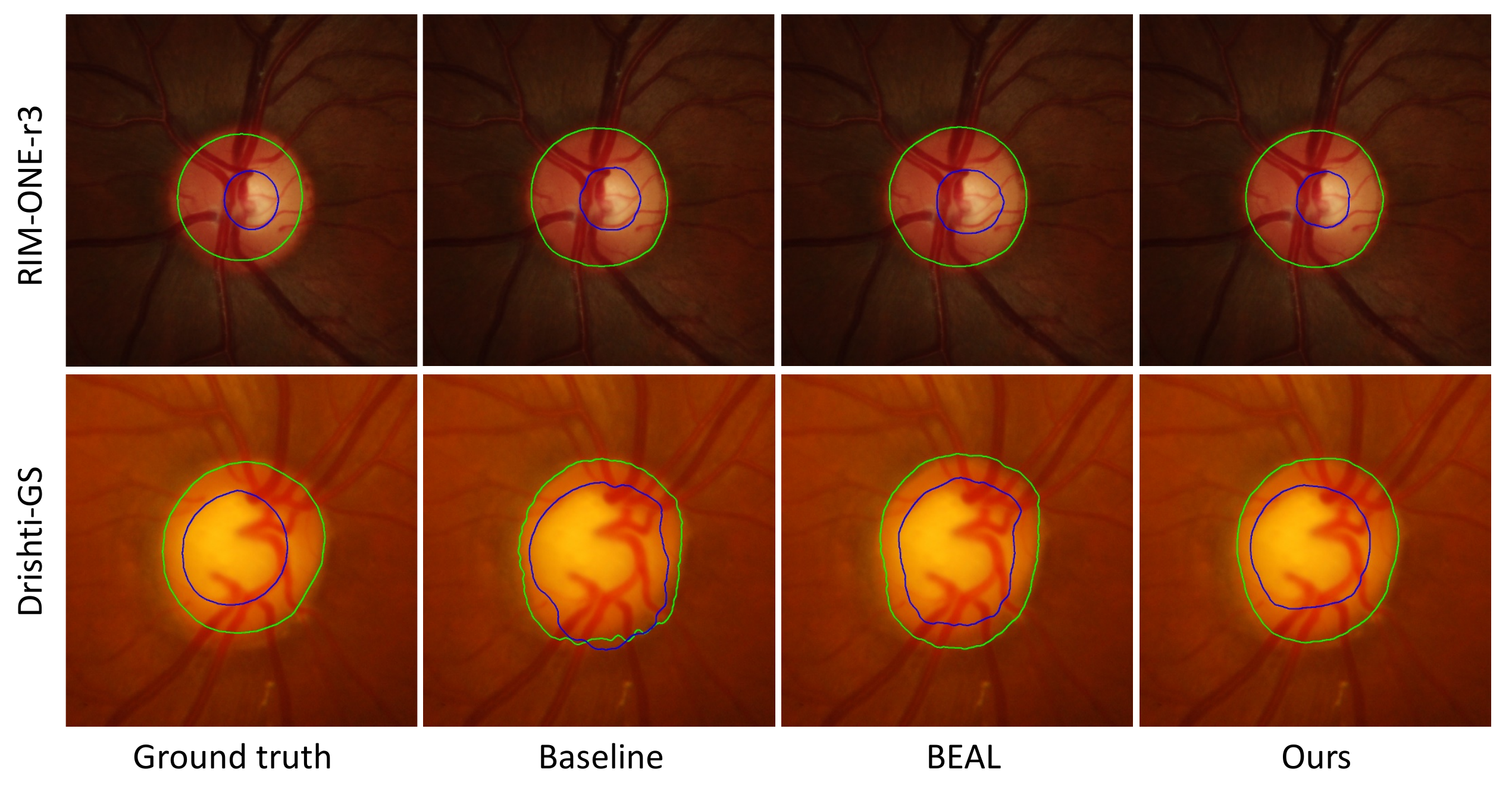} 

    \caption{Quantitative comparison of segmentation results of different methods}
    \label{fig:seg}

\end{figure*}

We also show the segmentation results of the different algorithms on the two datasets in \figref{fig:seg}. It can be seen that in some regions that are obscured or blurred by blood vessels, the segmentation results of other comparison algorithms are poor, while our method is able to accurately identify the boundaries of the optic cup and optic disc, while being very close to the ground truth labels.
\subsubsection{Ablation study.}
We conduct ablation experiments to investigate the effectiveness of each
component of the proposed method. In \tabref{fig_ab}, +src\_reg represents source domain category regularization, +trg\_reg denotes target domain category regularization, and +inter\_reg represents inter-domain category regularization. As seen in \tabref{fig_ab},
inter-domain category regularization and both intra-domain regularization
techniques lead to performance gains, which justifies the need for performing
global distribution regularization and category regularization simultaneously.
In addition, from \tabref{table_ab} we can also observe that using uncertainty-guided noise-aware (UGNA) modules
to filter out unreliable pseudo-labels can benefit inter-domain category
distribution regularization. By combining multiple category regularization
techniques, our approach further improves segmentation performance on both
datasets.
\begin{table*}[t!]
\centering
    \caption{Ablation study of different components of our method}
     \label{fig_ab}
\begin{tabular}{cccc|cccc}
\hline
\multicolumn{4}{l|}{\multirow{2}{*}{Method}}                                                               & \multicolumn{4}{c}{Target domain}                                                                         \\ \cline{5-8} 
\multicolumn{4}{l|}{}                                                                                      & \multicolumn{2}{c|}{RIM-ONE-r3\cite{fumero2011rim}}                                & \multicolumn{2}{c}{Drishti-GS\cite{sivaswamy2015comprehensive}}           \\ \hline
\multicolumn{1}{c|}{baseline} & \multicolumn{1}{c|}{+src\_reg} & \multicolumn{1}{c|}{+trg\_reg} & +inter\_reg & \multicolumn{1}{c|}{Dice disc} & \multicolumn{1}{c|}{Dice cup} & \multicolumn{1}{c|}{Dice disc} & Dice cup \\ \hline
\multicolumn{1}{c|}{\Checkmark}        & \multicolumn{1}{c|}{}         & \multicolumn{1}{c|}{}         &            & \multicolumn{1}{c|}{0.779}     & \multicolumn{1}{c|}{0.744}    & \multicolumn{1}{c|}{0.944}     & 0.836    \\ \hline
\multicolumn{1}{c|}{\Checkmark}        & \multicolumn{1}{c|}{\Checkmark}        & \multicolumn{1}{c|}{}         &            & \multicolumn{1}{c|}{0.899}     & \multicolumn{1}{c|}{0.829}    & \multicolumn{1}{c|}{0.958}     & 0.871    \\ \hline
\multicolumn{1}{c|}{\Checkmark}        & \multicolumn{1}{c|}{}         & \multicolumn{1}{c|}{\Checkmark}        &            & \multicolumn{1}{c|}{0.898}     & \multicolumn{1}{c|}{0.837}    & \multicolumn{1}{c|}{0.963}     & 0.875    \\ \hline
\multicolumn{1}{c|}{\Checkmark}        & \multicolumn{1}{c|}{}         & \multicolumn{1}{c|}{}         & \Checkmark          & \multicolumn{1}{c|}{0.901}     & \multicolumn{1}{c|}{0.833}    & \multicolumn{1}{c|}{0.961}     & 0.881    \\ \hline
\multicolumn{1}{c|}{\Checkmark}        & \multicolumn{1}{c|}{\Checkmark}        & \multicolumn{1}{c|}{\Checkmark}        &            & \multicolumn{1}{c|}{0.900}     & \multicolumn{1}{c|}{0.839}    & \multicolumn{1}{c|}{0.965}     & 0.880    \\ \hline
\multicolumn{1}{c|}{\Checkmark}        & \multicolumn{1}{c|}{\Checkmark}        & \multicolumn{1}{c|}{}         & \Checkmark          & \multicolumn{1}{c|}{0.903}     & \multicolumn{1}{c|}{0.836}    & \multicolumn{1}{c|}{0.964}     & 0.883    \\ \hline
\multicolumn{1}{c|}{\Checkmark}        & \multicolumn{1}{c|}{}         & \multicolumn{1}{c|}{\Checkmark}        & \Checkmark          & \multicolumn{1}{c|}{0.902}     & \multicolumn{1}{c|}{0.838}    & \multicolumn{1}{c|}{0.963}     & 0.887    \\ \hline
\multicolumn{1}{c|}{\Checkmark}        & \multicolumn{1}{c|}{\Checkmark}        & \multicolumn{1}{c|}{\Checkmark}        & \Checkmark          & \multicolumn{1}{c|}{\textbf{0.905}}     & \multicolumn{1}{c|}{\textbf{0.841}}    & \multicolumn{1}{c|}{\textbf{0.966}}     & \textbf{0.892}    \\ \hline
\end{tabular}

\end{table*}
\begin{table}[t!]
\centering
        \caption{The impact of UGNA in inter-domain category regularization} 
    \label{table_ab} 
\begin{tabular}{l|cc|cc}
\hline
\multirow{2}{*}{Method} & \multicolumn{2}{c|}{RIM-ONE-r3 \cite{fumero2011rim}}   & \multicolumn{2}{c}{Drishti-GS \cite{sivaswamy2015comprehensive}}   \\ \cline{2-5} 
                        & \multicolumn{1}{c|}{Dice disc} & Dice cup & \multicolumn{1}{c|}{Dice disc} & Dice cup \\ \hline
+inter\_reg(W/o UGNA)     & \multicolumn{1}{c|}{0.898}     & 0.824    & \multicolumn{1}{c|}{0.959}     & 0.870    \\ \hline
+inter\_reg     & \multicolumn{1}{c|}{\textbf{0.901}}     & \textbf{0.833}    & \multicolumn{1}{c|}{\textbf{0.961}}     & \textbf{0.881}    \\ \hline
\end{tabular}
\end{table}

\section{Conclusion}
In this paper, we propose an unsupervised domain adaptation framework based on category-level regularization for cross-domain fundus image segmentation. Three category regularization methods are developed to simultaneously regularize the category distribution from three perspectives: inter-domain, source and target domains, thus making the model better adapted to the target domain. Our method significantly outperforms state-of-the-art comparison algorithms on two public fundus datasets, demonstrating its effectiveness, and it can be applied to other unsupervised domain adaptation tasks as well.

%
%
%
\bibliographystyle{splncs04}
\bibliography{Paper_139}

\begin{thebibliography}{10}
\providecommand{\url}[1]{\texttt{#1}}
\providecommand{\urlprefix}{URL }
\providecommand{\doi}[1]{https://doi.org/#1}

\bibitem{chen2021source}
Chen, C., Liu, Q., Jin, Y., Dou, Q., Heng, P.A.: Source-free domain adaptive
  fundus image segmentation with denoised pseudo-labeling. In: International
  Conference on Medical Image Computing and Computer-Assisted Intervention. pp.
  225--235. Springer (2021)

\bibitem{chen2018encoder}
Chen, L.C., Zhu, Y., Papandreou, G., Schroff, F., Adam, H.: Encoder-decoder
  with atrous separable convolution for semantic image segmentation. In:
  Proceedings of the European conference on computer vision (ECCV). pp.
  801--818 (2018)

\bibitem{chen2020simple}
Chen, T., Kornblith, S., Norouzi, M., Hinton, G.: A simple framework for
  contrastive learning of visual representations. In: International conference
  on machine learning. pp. 1597--1607. PMLR (2020)

\bibitem{dou2018unsupervised}
Dou, Q., Ouyang, C., Chen, C., Chen, H., Heng, P.A.: Unsupervised
  cross-modality domain adaptation of convnets for biomedical image
  segmentations with adversarial loss. arXiv preprint arXiv:1804.10916  (2018)

\bibitem{fumero2011rim}
Fumero, F., Alay{\'o}n, S., Sanchez, J.L., Sigut, J., Gonzalez-Hernandez, M.:
  Rim-one: An open retinal image database for optic nerve evaluation. In: 2011
  24th international symposium on computer-based medical systems (CBMS).
  pp.~1--6. IEEE (2011)

\bibitem{gal2016dropout}
Gal, Y., Ghahramani, Z.: Dropout as a bayesian approximation: Representing
  model uncertainty in deep learning. In: international conference on machine
  learning. pp. 1050--1059. PMLR (2016)

\bibitem{hang2020local}
Hang, W., Feng, W., Liang, S., Yu, L., Wang, Q., Choi, K.S., Qin, J.: Local and
  global structure-aware entropy regularized mean teacher model for 3d left
  atrium segmentation. In: International Conference on Medical Image Computing
  and Computer-Assisted Intervention. pp. 562--571. Springer (2020)

\bibitem{hoffman2016fcns}
Hoffman, J., Wang, D., Yu, F., Darrell, T.: Fcns in the wild: Pixel-level
  adversarial and constraint-based adaptation. arXiv preprint arXiv:1612.02649
  (2016)

\bibitem{javanmardi2018domain}
Javanmardi, M., Tasdizen, T.: Domain adaptation for biomedical image
  segmentation using adversarial training. In: 2018 IEEE 15th International
  Symposium on Biomedical Imaging (ISBI 2018). pp. 554--558. IEEE (2018)

\bibitem{ju2022improving}
Ju, L., Wang, X., Wang, L., Mahapatra, D., Zhao, X., Zhou, Q., Liu, T., Ge, Z.:
  Improving medical images classification with label noise using
  dual-uncertainty estimation. IEEE transactions on medical imaging  (2022)

\bibitem{liu2021bapa}
Liu, Y., Deng, J., Gao, X., Li, W., Duan, L.: Bapa-net: Boundary adaptation and
  prototype alignment for cross-domain semantic segmentation. In: Proceedings
  of the IEEE/CVF International Conference on Computer Vision. pp. 8801--8811
  (2021)

\bibitem{orlando2020refuge}
Orlando, J.I., Fu, H., Breda, J.B., van Keer, K., Bathula, D.R., Diaz-Pinto,
  A., Fang, R., Heng, P.A., Kim, J., Lee, J., et~al.: Refuge challenge: A
  unified framework for evaluating automated methods for glaucoma assessment
  from fundus photographs. Medical image analysis  \textbf{59},  101570 (2020)

\bibitem{sivaswamy2015comprehensive}
Sivaswamy, J., Krishnadas, S., Chakravarty, A., Joshi, G., Tabish, A.S.,
  et~al.: A comprehensive retinal image dataset for the assessment of glaucoma
  from the optic nerve head analysis. JSM Biomedical Imaging Data Papers
  \textbf{2}(1), ~1004 (2015)

\bibitem{sohn2020fixmatch}
Sohn, K., Berthelot, D., Carlini, N., Zhang, Z., Zhang, H., Raffel, C.A.,
  Cubuk, E.D., Kurakin, A., Li, C.L.: Fixmatch: Simplifying semi-supervised
  learning with consistency and confidence. Advances in Neural Information
  Processing Systems  \textbf{33},  596--608 (2020)

\bibitem{wang2019boundary}
Wang, S., Yu, L., Li, K., Yang, X., Fu, C.W., Heng, P.A.: Boundary and
  entropy-driven adversarial learning for fundus image segmentation. In:
  International Conference on Medical Image Computing and Computer-Assisted
  Intervention. pp. 102--110. Springer (2019)

\bibitem{wang2019patch}
Wang, S., Yu, L., Yang, X., Fu, C.W., Heng, P.A.: Patch-based output space
  adversarial learning for joint optic disc and cup segmentation. IEEE
  transactions on medical imaging  \textbf{38}(11),  2485--2495 (2019)

\bibitem{wang2021deep}
Wang, X., Chen, H., Xiang, H., Lin, H., Lin, X., Heng, P.A.: Deep virtual
  adversarial self-training with consistency regularization for semi-supervised
  medical image classification. Medical image analysis  \textbf{70},  102010
  (2021)

\bibitem{wu2018multiscale}
Wu, Y., Xia, Y., Song, Y., Zhang, Y., Cai, W.: Multiscale network followed
  network model for retinal vessel segmentation. In: International Conference
  on Medical Image Computing and Computer-Assisted Intervention. pp. 119--126.
  Springer (2018)

\bibitem{xie2018learning}
Xie, S., Zheng, Z., Chen, L., Chen, C.: Learning semantic representations for
  unsupervised domain adaptation. In: International conference on machine
  learning. pp. 5423--5432. PMLR (2018)

\bibitem{yu2019uncertainty}
Yu, L., Wang, S., Li, X., Fu, C.W., Heng, P.A.: Uncertainty-aware
  self-ensembling model for semi-supervised 3d left atrium segmentation. In:
  International Conference on Medical Image Computing and Computer-Assisted
  Intervention. pp. 605--613. Springer (2019)

\bibitem{zhang2018task}
Zhang, Y., Miao, S., Mansi, T., Liao, R.: Task driven generative modeling for
  unsupervised domain adaptation: Application to x-ray image segmentation. In:
  International Conference on Medical Image Computing and Computer-Assisted
  Intervention. pp. 599--607. Springer (2018)

\bibitem{zheng2020cross}
Zheng, Y., Huang, D., Liu, S., Wang, Y.: Cross-domain object detection through
  coarse-to-fine feature adaptation. In: Proceedings of the IEEE/CVF conference
  on computer vision and pattern recognition. pp. 13766--13775 (2020)

\bibitem{zhu2017unpaired}
Zhu, J.Y., Park, T., Isola, P., Efros, A.A.: Unpaired image-to-image
  translation using cycle-consistent adversarial networks. In: Proceedings of
  the IEEE international conference on computer vision. pp. 2223--2232 (2017)

\end{thebibliography}
%





\end{document}